\documentclass[11pt]{article}
\usepackage{geometry}
\usepackage{coling2020}
\usepackage{times}
\usepackage{url}
\usepackage{latexsym}
\usepackage{microtype}
\usepackage{graphicx}
\usepackage{MnSymbol,wasysym}
\usepackage{soul}
\usepackage{svg}

\colingfinalcopy 

\title{LIMSI\_UPV at SemEval-2020 Task 9: Recurrent Convolutional Neural Network for Code-mixed Sentiment Analysis}

\author{
  Somnath Banerjee$^1$, Sahar Ghannay$^1$, Sophie Rosset$^1$, Anne Vilnat$^1$, Paolo Rosso$^2$ \\
  $^1$ LIMSI, CNRS, Universit\'e Paris-Saclay, Orsay, France\\
  $^2$ Universitat Polit\`{e}cnica de Val\`{e}ncia, Spain \\
  $^1${\tt firstname.lastname@limsi.fr}\\
  $^2${\tt prosso@dsic.upv.es}}

\date{}

\begin{document}
\maketitle
\begin{abstract}
  This paper describes the participation of LIMSI\_UPV team in SemEval-2020 Task 9: Sentiment Analysis for Code-Mixed Social Media Text.
The proposed approach competed in SentiMix Hindi-English subtask, that addresses the problem of predicting the sentiment of a given Hindi-English code-mixed tweet. 
We propose Recurrent Convolutional Neural Network that combines both the recurrent neural network and the convolutional network to better capture the semantics of the text, for code-mixed sentiment analysis. 
The proposed system obtained 0.69 (best run) in terms of F1 score on the given test data and achieved the 9th place (Codalab username: somban) in the SentiMix Hindi-English subtask.
\end{abstract}

\section{Introduction}
\label{intro}

%
%
\blfootnote{
    %
    %
    \hspace{-0.65cm}  
    %
    %
     This work is licensed under a Creative Commons Attribution 4.0 International Licence.
     Licence details: \url{http://creativecommons.org/licenses/by/4.0/}.
    %
    %
}
\blfootnote{The code of this work is available at \url{https://github.com/somnath-banerjee/Code-Mixed_SentimentAnalysis}.}

 In this digital era, users express their personal thoughts and opinions regarding a wide range of topics on social media platforms such as blogs, micro-blogs (e.g., Twitter), and chats (e.g., WhatsApp and Facebook messages). 
 Multilingual societies like India with a decent amount of internet penetration widely adopted such social media platforms. 
 However, the regional language influences the proliferation of the Hindi-English Code-Mixed (CM) data. 
 Sentiment analysis of these end-user data from social media is a crucial resource for commerce and governance.
 However, in contrast to the classical sentiment analysis methods, which were originally designed for dealing with well-written product reviews, CM texts from social media often contain misspellings (often intentional), badly cased words, letter substitutions, ambiguities, non standard abbreviations, improper use of grammar, etc.
 
 CM poses several unseen difficulties to natural language processing (NLP) tasks such as word-level language identification, part-of-speech tagging, dependency parsing, machine translation and semantic processing. 
 In the last few years, a number of workshops such as \textit{Linguistic Code-Switching Workshops}\footnote{https://code-switching.github.io/2020/} and shared tasks such as \textit{Mixed Script Information Retrieval} \cite{banerjee:msir}  have been organized due to the emerging popularity of code-mixing.   
 To promote research in this area, Task 9 of SemEval-2020 was devoted to CM sentiment analysis in Twitter. 
 The goal of the task was to automatically classify the polarity of a given CM Twitter post into one of the three predefined categories: \textit{positive}, \textit{negative} and \textit{neutral}. The CM languages are English-Hindi and English-Spanish; for a more detailed description of the task see \cite{patwa2020sentimix}.  
 
In this paper, we present a deep learning approach, using a Recurrent Convolutional Neural Network for the task of automatic CM sentiment classification of tweets. 

The rest of the paper is structured as follows. 
Section 2 provides background in brief. 
Section 3 provides the system overview and Section 4 describes our approach in detail.
In Section 5, we discuss the analysis and evaluation results for our system.
We conclude our work in Section 6.

\section{Background}
\label{relwork}
Sentiment classification is the task of detecting whether a textual item (e.g., a product review, a blog post, an editorial, etc.) expresses a POSITIVE or a NEGATIVE opinion in general or about a given entity, e.g., a product, a person, a political party, or a policy \cite{nakov-semeval:2016}.
Classifying tweets according to sentiment has many applications in political science, social sciences, market research, and many others \cite{martinez:2014,mejova:2015}.
Although initially sentiment identification was focused on newswire text \cite{baccianella:2010}, later research turned towards social media \cite{semeval:2015}.
Since 2013, a sentiment classification task on Twitter data have been organized in SemEval campaigns.

Most of the earlier approaches to this problem were based on hand crafted features and sentiment lexicons \cite{pak-paroubek:2010,mohammad-et-al:2013}. These features were then used as input to classifiers (e.g., Support Vector Machines). However, such approaches required extensive domain knowledge, were laborious to define, and can lead to incomplete or over-specific features.

Recently, researchers pay their attention to sentiment polarity detection on CM data. However, a few research work have been carried out in particular Hindi-English CM data with different approaches: lexicon lookup \cite{sharma-et-al:2015}, sub-word with CNN-LSTM \cite{joshi-et-al:2016}, Siamese networks \cite{choudhary-et-al:2018}, dual Encoder Network with features \cite{lal-et-al:2019}.  
Lai et al. \shortcite{lai:2015} proposed Recurrent Convolutional Network for text classification which is a foundational task in many NLP applications. We followed this model in our task.

\section{System overview}
\label{sysovr}

We are inspired by the model proposed in \cite{lai:2015} particularly proposed for the text classification task.   
The proposed model takes sequence of CM words as input and provides sentiment polarity class as output. 
The recurrent structure of the proposed model captures the contextual information during the learning of the word representation, and the max-pooling layer identifies the key CM words.  
If $T$, $s$ and $\theta$ denote a CM tweet made up of sequence of CM words (T = {$w_1$, $w_2$, $w_3$, \dots, $w_n$}), the sentiment polarity class and parameters of the neural network respectively, $p(s|T,\theta))$ denotes the probability of the tweet $T$ having sentiment polarity $s$, where $s$ 
could be any one of sentiment polarity classes, i.e., $s\in$\{\textit{positive, negative, neutral}\}.

In this model, a CM word is represented by combining the word and its both side contexts. If $c_l(w_i)$ and $c_r(w_i)$ denote the left-side and right-side contexts of a CM word $w_i$, a word is represented as a concatenation of the left-side context vector $c_l(w_i)$, the word embedding $e(w_i)$ and the right-side context vector $c_r(w_i)$. 
  \[x_i = [c_l(w_i); e(w_i); c_r(w_i)]\]

where, $x_i$ is the representation of the $i^{th}$ word (i.e., $w_i$) in $T$.

Both the contexts (i.e., $c_l(w_i)$ and $c_r(w_i)$) are calculated as follows:

\[c_l(w_i) = f(W^{(l)}c_l(w_{i-1}) + W^{(s_l)}e(w_{i-1}))\]
\[c_r(w_i) = f(W^{(r)}c_r(w_{i+1}) + W^{(s_r)}e(w_{i+1}))\] 

Where, 

\qquad \quad $f$: is a non-linear activation function;

\qquad \quad $e(w_{i-1})$: the word embedding of word $w_{i-1}$; 

\qquad \quad $W^{(l)}$: a matrix that transforms the hidden layer (context) into the next hidden layer;

\qquad \quad $W^{(sl)}$: a matrix that is used to combine the semantic of the current word with the next word's left context.
     
In the forward scan, the recurrent structure of the model obtains all the $c_l$ of the CM tweet, whereas, it obtains all the $c_r$ in a backward scan of the CM tweet. 
After obtaining the $x_i$ for the word $w_i$, a linear transformation together with the tanh activation function is applied to $x_i$ and the result is sent to the next layer:

\[y^{(2)}_i = \textrm{tanh} (W^{(2)}x_i + b^{(2)})\]

After calculating all of the representations of words, a max-pooling layer is applied. 
Hence, the pooling layer utilizes the output of the recurrent structure as the input. 
This layer attempts to find the most important latent semantic factors in the CM tweet:

\[y^{(3)} = \max^{n}_{i=1} y^{(2)}_i\]

Finally, in the output layer, a softmax function is applied that provides the sentiment polarity probabilities:
\[y^{(4)} = W^{(4)}y^{(3)} + b^{(4)}\]
\[p_i = \frac{\textrm{exp}(y_{k}^{(4)})}{\displaystyle\sum_{k=1}^{n} \textrm{exp}(y_{k}^{(4)})}\]

\begin{figure}[h]
\centering
\includegraphics[width=\linewidth]{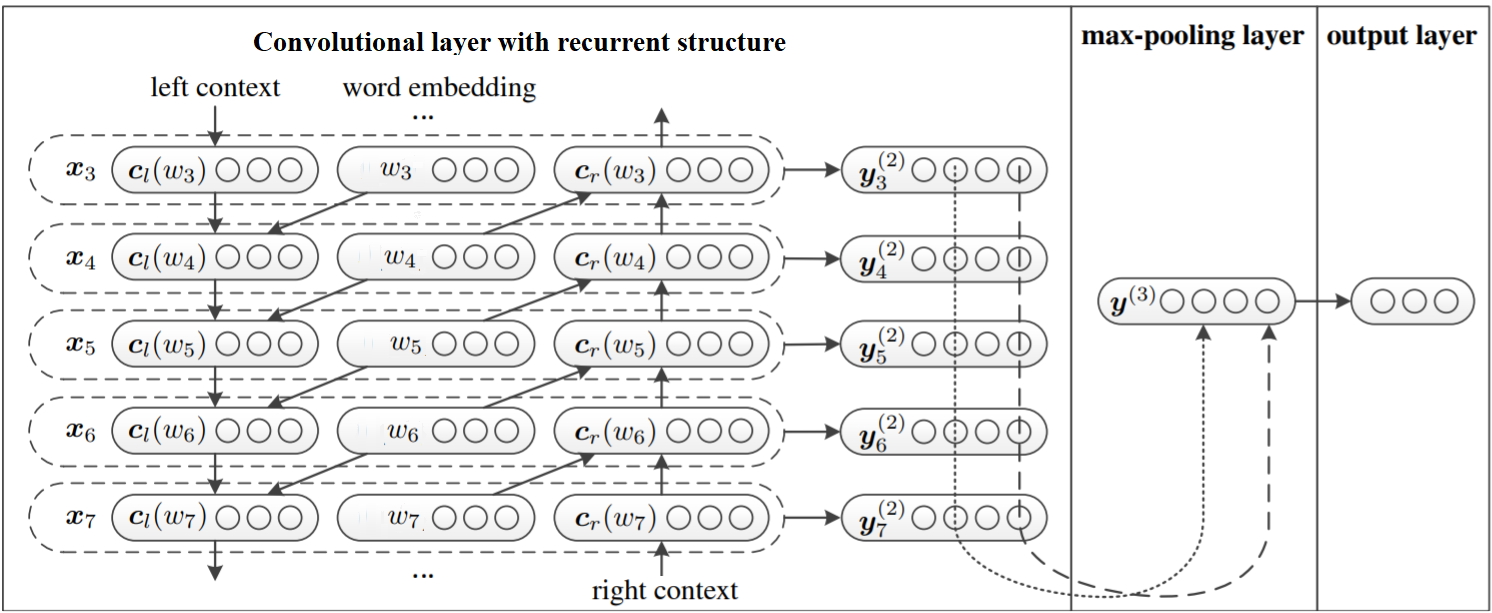}
\caption{Recurrent Convolutional Neural Network \cite{lai:2015} model}
\label{fig:model}
\end{figure}

The Figure~\ref{fig:model} depicts the architecture of the proposed model. From Figure~\ref{fig:model}, we could see that $c_l(w_n)$ captures the left-side context, whereas, $c_l(w_1)$ captures the right-side context of a word $w_i$.

\section{Experimental setup}
\label{expsetup}

\textbf{Data:} For sentiment analysis on Hindi-English CM tweets, we used the dataset provided by the organizers of Task 9 at SemEval-2020. 
The training dataset consists of 14 thousand tweets. 
Whereas, the validation dataset as well as the test dataset contain 3 thousand tweets each. 
The details of the dataset are given in \cite{patwa2020sentimix}.
For this task, we did not use any external dataset.

\textbf{Preprocessing:} 
The CM Twitter data is not formal in nature. 
Also, often casing is not followed properly.
As a result, it misleads the language processing task.
Therefore, we converted the entire tweets into lower case. 
After observing the CM dataset, we found that the urls are not contributing any information. 
While writing, the writer of a tweet addresses a person by mentioning the person's name followed by @ or by twitter id which starts with @. 
Similarly, to specify a particular topic, the writer uses the topic name that starts with \#. 
To capture the semantics and syntax of a tweet, we replaced the person and a topic being addressed with a MENTION and a TOPIC token respectively. 
The following steps were applied on each CM tweet:
\begin{itemize}
  \item Tokens were converted to lowercase.
  
  \item If a token is http or https, the following tokens are merged to identify as links and were deleted. 
  
  \item Garbage tokens (e.g., â€¦, $\grave{a}$ ¥¥, etc.)
  were deleted.
  
  \item If a token is @, the following token is merged with the first token (i.e., @) and the new token is replaced with a MENTION token. For example, the tokens @ and \textit{bomanirani} are merged and replaced with a MENTION token.
  
  \item If a token is \#, the following token is merged with the first token (i.e., \#) and the new token is replaced with a TOPIC token. 
  For example, the tokens \# and \textit{LoveIsLove} are merged and replaced with a TOPIC token.
  
  \item Emoji's with text were divided into two tokens. For example, \textit{he}\smiley{} becomes \textit{he} and \smiley{}.
  
  \item If a token contains more than one emoji, each emoji was considered as a token. For example, \smiley{}\frownie{} becomes \smiley{} and \frownie{}.
\end{itemize}

\textbf{Embeddings:}  Following Collobert et al. \shortcite{collobert-et-al:2011}, a lot of authors 
argued that word embedding plays a vital role to improve natural language task performance. Hence, we experimented the use of word embeddings to improve the performance of our proposed models.
Using the fastText \cite{fasttext}, we prepared two embedding models: Skip-gram and Cbow.
After empirically evaluating the performance on validation set, the embeddings` dimensionality was set to 300 for all the embeddings. 
The embeddings are trained on training data using the parameters: lr=0.05, context window=5, epochs=5, minimal number of word occurences=5, dimensionality=300.  

\textbf{Experiment:} We carried out two experiments with similar settings except different word embedding approaches:  Skip-gram for SkipGRun, and Cbow for CbowRun.

\textbf{Hyper-parameters:} After evaluating the model performance on the validation data, the optimal values of the hyper-parameters were set.
We used the following list of hyper-parameters: learning rate = 0.6, word embedding vector size = 300, hidden layers= 2, hidden layer size = 64, context vector size = 5, dropout rate = 0.1, optimizer = stochastic gradient descent, loss function= negative log likelihood, and batch size = 64. 

\section{Results and Analysis}
\label{res}
As mentioned in the previous section, we carried out experiments with our proposed model using two settings. 
All the presented experiments are evaluated on the test data for the given task.
The performance of the systems were evaluated using F1 averaged across the positive, negative and the neutral. 

\begin{table}[h]
\centering
\begin{tabular}{lrrr}
                               & \multicolumn{1}{l}{Precision} & \multicolumn{1}{l}{Recall} & \multicolumn{1}{l}{F1 score} \\ \hline\hline
\multicolumn{1}{l}{Baseline} & \multicolumn{1}{r}{-}         & \multicolumn{1}{r}{-}      & \multicolumn{1}{r}{0.654}   \\ \hline
\multicolumn{1}{l}{SkipGRun}    & \multicolumn{1}{r}{0.6952}       & \multicolumn{1}{r}{0.6893}   & \multicolumn{1}{r}{\textbf{0.6913}} \\   
\multicolumn{1}{l}{CbowRun}    & \multicolumn{1}{r}{0.6566}      & \multicolumn{1}{r}{0.6556}   & \multicolumn{1}{r}{0.6560}  
\\ \hline\hline
\end{tabular}
\caption{Evaluation results on test data}
\label{tab:test_results}
\end{table}

The organizer baseline F1 scores for the validation and test data are 0.58 and 0.654 respectively.
The details of the baseline are given in \cite{patwa2020sentimix}. 
The obtained results with our submitted runs are given in Table~\ref{tab:test_results}.
For SkipGRun, we achieved 0.6913 of F1 score with 0.6952 and 0.6893 of precision and recall respectively.
The SkipGRun outperformed the CbowRun by around 0.40 in terms of F1 score. 
CbowRun outperformed the organizers' baseline by a slight margin, however, SkipGRun outperformed the baseline by around 0.4 in terms of averaged F score.

For SkipGRun, the confusion matrix and the performance of class-wise sentiment polarities are presented 
in Table~\ref{tab:run-1:conf} and Table~\ref{tab:run-1:classwise}, respectively.
The SkipGRun is the best among the two submitted runs. 
From the confusion matrix (cf. Table~\ref{tab:run-1:conf}), we can observe that the system is more successful in identifying the \textit{positive} polarity class (F1 score 0.76) than the polarity classes: \textit{negative} (F1 score 0.71) and \textit{neutral} (F1 score 0.62).  
This run performed well on distinguishing a \textit{positive} tweet from \textit{negative} tweet and vice versa.
Therefore, only 35 \textit{positive} tweets were misclassified as \textit{negative} and 24 \textit{negative} tweets were misclassified as \textit{positive}. 
However, it has a problem to separate \textit{positive} and \textit{negative} tweets from \textit{neutral}. 
Hence, a lot of \textit{positive} and \textit{negative} tweets were misclassified as \textit{neutral}. 
From Table~\ref{tab:run-1:conf}, we could say that around 24\% of \textit{positive} tweets and 28\% of \textit{negative} tweets were identified as \textit{neutral}. 
Similarly, a number of \textit{neutral} tweets were misclassified as \textit{positive} and \textit{negative}.     
From Table~\ref{tab:run-1:classwise}, it is evident that the positive tweets were the easiest class to predict. The F1 scores were 0.76 (positive class), 0.71 (negative class) and 0.62 (neutral class).

\begin{table}[h]
\parbox{.5\linewidth}{
\centering
\begin{tabular}{lrrr}
                               & \multicolumn{1}{l}{Positive} & \multicolumn{1}{l}{Negative} & \multicolumn{1}{l}{Neutral} \\ \hline \hline
\multicolumn{1}{l}{Positive} & \multicolumn{1}{r}{729}     & \multicolumn{1}{r}{35}      & \multicolumn{1}{r}{236}    \\ \hline
\multicolumn{1}{l}{Negative} & \multicolumn{1}{r}{24}      & \multicolumn{1}{r}{624}     & \multicolumn{1}{r}{252}    \\ \hline
\multicolumn{1}{l}{Neutral}  & \multicolumn{1}{r}{175}     & \multicolumn{1}{r}{210}     & \multicolumn{1}{r}{715}    \\ \hline\hline
\end{tabular}
\caption{SkipGRun: Confusion matrix}
\label{tab:run-1:conf}
}
\hfill
\parbox{.5\linewidth}{
\centering
\begin{tabular}{lrrr}
                               & \multicolumn{1}{l}{Precision} & \multicolumn{1}{l}{Recall} & \multicolumn{1}{l}{F1 score} \\ \hline \hline
\multicolumn{1}{l}{Positive} & \multicolumn{1}{r}{0.79}     & \multicolumn{1}{r}{0.73}      & \multicolumn{1}{r}{0.76}    \\ \hline
\multicolumn{1}{l}{Negative} & \multicolumn{1}{r}{0.72}      & \multicolumn{1}{r}{0.69}     & \multicolumn{1}{r}{0.71}    \\ \hline
\multicolumn{1}{l}{Neutral}  & \multicolumn{1}{r}{0.59}     & \multicolumn{1}{r}{0.65}     & \multicolumn{1}{r}{0.62}    \\ \hline\hline
\end{tabular}
\caption{SkipGRun: Polarity class-wise performance}
\label{tab:run-1:classwise}
}
\end{table}

Table~\ref{tab:run-2:conf} and Table~\ref{tab:run-2:classwise} present the confusion matrix and the performance of class-wise sentiment polarities for CbowRun.
Like SkipGRun, the CbowRun is also able to successfully identify the \textit{positive} polarity class (F1 score 0.72) in comparison to others (cf. Table~\ref{tab:run-2:conf}).
For CbowRun, we observed similar characteristics of results like SkipGRun. 
Although the results obtained for \textit{negative} tweets are almost similar with SkipGRun (F1 score 0.71) and CbowRun (F1 score 0.70), the difference in performance is notable for \textit{positive} and \textit{neutral} tweets.  
From Table~\ref{tab:run-2:classwise}, we can see that for CbowRun the F1 scores were 0.73 (positive class), 0.70 (negative class) and 0.57 (neutral class).

\begin{table}[h]
\parbox{.5\linewidth}{
\centering
\begin{tabular}{lrrr}
                               & \multicolumn{1}{l}{Positive} & \multicolumn{1}{l}{Negative} & \multicolumn{1}{l}{Neutral} \\ \hline \hline
\multicolumn{1}{l}{Positive} & \multicolumn{1}{r}{709}     & \multicolumn{1}{r}{41}      & \multicolumn{1}{r}{250}    \\ \hline
\multicolumn{1}{l}{Negative} & \multicolumn{1}{r}{33}      & \multicolumn{1}{r}{630}     & \multicolumn{1}{r}{237}    \\ \hline
\multicolumn{1}{l}{Neutral}  & \multicolumn{1}{r}{232}     & \multicolumn{1}{r}{240}     & \multicolumn{1}{r}{628}    \\ \hline\hline
\end{tabular}
\caption{CbowRun: Confusion matrix}
\label{tab:run-2:conf}
}
\hfill
\parbox{.5\linewidth}{
\centering
\begin{tabular}{lrrr}
                               & \multicolumn{1}{l}{Precision} & \multicolumn{1}{l}{Recall} & \multicolumn{1}{l}{F1 score} \\ \hline \hline
\multicolumn{1}{l}{Positive} & \multicolumn{1}{r}{0.73}     & \multicolumn{1}{r}{0.71}      & \multicolumn{1}{r}{0.72}    \\ \hline
\multicolumn{1}{l}{Negative} & \multicolumn{1}{r}{0.69}      & \multicolumn{1}{r}{0.70}     & \multicolumn{1}{r}{0.70}    \\ \hline
\multicolumn{1}{l}{Neutral}  & \multicolumn{1}{r}{0.56}     & \multicolumn{1}{r}{0.57}     & \multicolumn{1}{r}{0.57}    \\ \hline\hline
\end{tabular}
\caption{CbowRun: Polarity class-wise performance}
\label{tab:run-2:classwise}
}
\end{table}

To get a deeper analysis of the results, we also performed the error analysis. 
We observed that the system could not identify the sentiment when Hindi lyrics are used in tweets. For example: \textit{RT MENTION Aankho ki hai ye khawaise ki chehre se teri na hate ...}  (id:36925, gold:\textit{positive}, predicted:\textit{negative}); \textit{RT MENTION Saare jahan se achha \# Hindustan hamara Ham bulbulain hai iski yeh gulsitan hamara...}  (id:32707, gold:\textit{positive},predicted:\textit{neutral}).


The system often failed to identify the sentiment, when a long tweet consists of a number of complete or incomplete sentence with mixed sentiments. For example: \textit{RT MENTION I miss childhood days ... No problems ... No hates ... No shames ... No stress ... No heartbreaks ... Go school ... Life was easy} (id:25011, gold:\textit{neutral} predicted:\textit{positive}); \textit{RT MENTION PM Modi won 356 seats \& thanked all 130 crore people for faith in democracy . Sonia won 52 seats \& thanked only the 12 cro }(id:30674, gold:\textit{positive},  predicted:\textit{neutral})

The proposed system often failed to identify the tweets that have 
any punctuation or delimiter to separate the sentence clauses. For example: \textit{Jb Koi aapke liye kam krta h aapke liye 5sal din rat ek krta h aap pe vishvas krta h aapka vishvas jit ta h jis} (id:31020, gold:\textit{neutral}, predicted:\textit{positive})

We observed that mostly the system misclassified the \textit{positive} and \textit{negative} tweets as \textit{neutral} and vice versa. However, we observed that there are some tweets that may arise some arguments, such as \textit{MENTION Beautiful words and its one of my fav song} (id:31662), \textit{MENTION Sir me from Bihar Love you You have Great job for Bjp continue We are with you} (id:39755), etc. tagged as \textit{neutral} and our system identified as \textit{positive}.
Identifying the polarity of an entity or entity’s aspect in the tweets along with more training data could help to resolve these issues.%

\section{Conclusion}
\label{conclusion}
This paper describes the approach we proposed for SemEval-2020 Task 9: Sentiment Analysis for CM Social Media Text (SentiMix Hindi-English). 
In our approach, we pre-processed the CM tweets and proposed a Recurrent Convolutional Neural Network for the sentiment analysis of CM tweets.
We submitted two runs and obtaining promising results: our best run obtained 0.691 of F1 averaged across the positives, negatives and the neutral. 
We observed that the proposed architecture occasionally strives to separate the \textit{positive} and \textit{negative} polarities from the \textit{neutral} and vice versa.  

For future work, we will explore the performance of our model with larger corpora against the
testing set. Also, we would like to investigate other embedding choices such as BERT \cite{bert:2019}. Moreover, due to the impact that irony and sarcasm have on sentiment analysis \cite{hernandez-rosso:2016} it would be
interesting to apply deep learning techniques to detect irony \cite{zhang-etal:2019} but in a code-mixed scenario.
    
\section*{Acknowledgements}
The research work of the first four authors was supported by ERA-Net CHIST-ERA LIHLITH Project funded by ANR (France) project ANR-17-CHR2-0001-03.
The research work of the last author was partially funded by the Spanish MICINN under the project MISMIS-FAKEnHATE on Misinformation and Miscommunication in social media: FAKE news and HATE speech (PGC2018-096212-B-C31).

\bibliographystyle{coling}
\bibliography{semeval2020}
\end{document}